\documentclass[11pt,a4paper]{article}
\usepackage[hyperref]{acl2017}
\usepackage{times}
\usepackage{latexsym}
\usepackage{graphicx}
\usepackage{breqn}

\usepackage{url}

\aclfinalcopy 

\title{A Survey of Distant Supervision Methods using PGMs}

\author{Gagan Madan \\
  2013ME10015 \\
  Indian Institue of Technology Delhi \\
  {\tt gaganm.me113@mech.iitd.ac.in} \\}

\date{}

\begin{document}
\maketitle
\begin{abstract}
Relation Extraction refers to the task of populating a database with tuples of the form r($e_1$, $e_2$), where r is a relation and $e_1, e_2$ are entities. Distant supervision is one such technique which tries to automatically generate training examples based on an existing KB such as Freebase. This paper is a survey of some of the techniques in distant supervision which primarily rely on Probabilistic Graphical Models (PGMs). 
\end{abstract}

\section{Introduction}

In the recent years, there has been an explosion of information on the web. As the size of data available continues to expand rapidly, this provides an interesting challenge as well as an opportunity to researchers in NLP. While managing such large amounts of data, and distilling the relevant information in a form that is manageable is a challenge, the large amount of data available provides an opportunity to do this.

One possible way to manage such large amounts of data is to populate a database with tuples of the form \verb!relation<entity1, entity2>!, extracted from text. Examples of such tuples could be \verb!capital-of(Delhi, India)!, \verb!PrimeMinister(Narendra Modi, India)!. This simple representation is particularly useful, as it allows answering queries on the text directly about a given relation, and entities. To extract tuples of this form is the key objective of Relational Extraction. Relational Extraction is a particularly well studied problem in the NLP domain, with many varied existing approaches to this problem. 

Most of the approaches can be classified as follows:
\begin{enumerate}
\item Bootstrapping Methods: These methods start with a small set of ``seed'' tuples, and iteratively generate patterns from the seed tuples, recognize more tuples to be utilized as seed tuples, thus bootstrapping into the complete relation table. This was introduced for the first time in DIPRE \cite{brin1998extracting}, and later extended in Snowball \cite{agichtein2000snowball}. Bootstrapping methods typically suffer from semantic drift, and often have poor precision as the number of iterations increases.
\item Supervised Methods: These methods treat the task of relational extraction as a supervised learning problem, and rely on the availability of extensive training data for extracting relations. Due to the reliance on training data, these methods usually extract tuples only from some given relations. Some interesting approaches in this domain have used tree kernels built on dependency parse trees \cite{culotta2004dependency}, training semi-CRFs \cite{sarawagi2004semi}, and using shortest path tree kernels \cite{bunescu2005shortest}.
\item Open Domain IE Methods: This is a self-supervised learning model, which utilises no training data, but relies on redundancy of information present in the corpus. This outputs a collection of all possible tuples of the form $(e_1, r, e_2)$ where $e_1, e_2$ are two entities related through a phrase r. The current state-of-the-art system in OpenIE is Open IE 4.0, which is a further improvement on Reverb \cite{fader2011identifying} and Ollie \cite{schmitz2012open}.
\item Distant Supervision: This technique automatically generates training examples and tries to learn features based on target relational tables in a Knowledge Base (KB). Typically, large KBs such as Freebase are used for this. This technique usually does not require any human intervention. This technique was originally introduced in the context of biological KBs \cite{craven1999constructing}, but has been successfully extended to any texts \cite{mintz2009distant}.
\item Deep Learning Based Methods: These techniques are relatively new and utilize the word embeddings generated by word2vec \cite{mikolov2013distributed}. One of the most successful atttempts in this domain, \cite{zeng2014relation} uses Convolution Neural Networks (CNNs) for relational extraction.
\end{enumerate}
While a major focus currently in NLP is on deep learning techniques, as KBs become better with time, it makes sense to focus on techniques that can leverage the power of KBs. With this in mind, it makes sense to focus on Distant Supervision, a technique which requires minimal human intervention and is primarily based on utilising large scale KBs such as Freebase.

\section{Problem Definition}
We define the task of distant supervision formally in the notation given by \cite{min2013distant}:\\
Given a knowledge base (KB), $\mathcal{D}$, a set of relations $R$, the KB $\mathcal{D}$ contains tuples of the form $r(e_1, e_2)$, where $r \in R$, and $e_1$ and $e_2$ are entities known to be related by the relation $r$. Further, we are given a corpus, $\mathcal{C}$, which contains natural language text. The task in Distant Supervision is to {\it align} the corpus $\mathcal{C}$ with the KB $\mathcal{D}$, i.e. to automatically generate training examples for relational extraction by labelling relational mentions in $\mathcal{C}$ with relations in $\mathcal{D}$. Formally, this requires labelling set of entity mentions $(e_1, e_2)$ present in $\mathcal{C}$ with some $r \in R$ or $OTHER$. Most works treat the $OTHER$ class as negative training examples.

\section{Datasets and Knowledge Bases}
\subsection{Datasets}
Some of the earliest work in Distant Supervision was motivated by research in the biomedical domain. Therefore, the first model by \cite{craven1999constructing} primarily relied on database from medical sources. Subsequently, evaluations have been performed on Wikipedia, NY Times corpus and more recently on the 2010 and 2011 KBP shared tasks \cite{ji2010overview, owczarzak2011overview}. Some features of the datasets are:

\begin{enumerate}
\item Yeast Protein Database(YPD) \cite{hodges1999yeast}, PubMed, MEDLINE: The work by \cite{craven1999constructing} primarily uses the YPD database, which includes facts about various proteins as well as links to PubMed articles that establish the fact. Further, they use MEDLINE, a database of bibliographic information and abstracts for over nine million articles in biomedical domain for evaluation of their model.
\item Freebase-Wikipedia: This dataset is a dump of all Wikipedia articles which have been sentence tokenized by Metaweb Technologies, the developers of Freebase. This is the main dataset used by \cite{mintz2009distant} for Distant Supervision.
\item NY Times: This dataset was developed by \cite{riedel2010modeling} by aligning Freebase relations with the NY Times dataset. They use StanfordNER to find relevant entity mentions in the text.
\item KBP: This dataset was primarily used for evaluation by \cite{surdeanu2012multi}. The training relations here are a were generated from the 2010 and 2011 KBP shared tasks \cite{ji2010overview, owczarzak2011overview}, which is a subset of Wikipedia Infoboxes from 2008.
\end{enumerate}

\subsection{Knowledge Bases}
Most Distant Supervision approaches typically use Freebase \cite{bollacker2008freebase} as the KB for the task. Freebase is a publicly available database of semantic data. It contains relations from various sources, primarily from Wikipedia text boxes. It has around 9 million entities and around 7300 relations. Some of the largest relations are shown in Table \ref{table:freebase}.

\begin{table*}[t]
\centering
\caption{10 largest Freebase relations \cite{mintz2009distant}}
\label{table:freebase}
\begin{tabular}{|l|l|l|}
\hline
Relation name                         & Size   & Example                                \\ \hline
/people/person/nationality            & 281107 & John Dugard, South Africa              \\
/location/location/contains           & 253223 & Belgium, Nijlen                        \\
/people/person/profession             & 208888 & Dusa McDuff, Mathematician             \\
/people/person/place of birth         & 105799 & Edwin Hubble, Marshfield               \\
/dining/restaurant/cuisine            & 86213  & MacAyo’s Mexican Kitchen, Mexican      \\
/business/business chain/location     & 66529  & Apple Inc., Apple Inc., South Park, NC \\
/biology/organism classification rank & 42806  & Scorpaeniformes, Order                 \\
/film/film/genre                      & 40658  & Where the Sidewalk Ends, Film noir     \\
/film/film/language                   & 31103  & Enter the Phoenix, Cantonese           \\ \hline
\end{tabular}
\end{table*}

\section{Distant Supervision using PGMs}
In this section, we describe the various models proposed for Distant Supervision using Probabilistic Graphical Models (PGMs). We begin our discussion with the simplest model, based on a Naive Bayes classifier, and move to models using more complicated PGMs.
\subsection{Distant Supervision in Biomedical Domain \cite{craven1999constructing}}
This paper was the first attempt at using an existing database to generate training examples from a text corpus. While the model itself was very simple, it was in many ways ahead of its time. Since there were no large scale general knowledge databases available around that time, while this model was used successfully in domain specific contexts such as genetics \cite{jenssen2001literature}, it wasn't until when large scale KBs such as Freebase became available that this technique started becoming relevant. 

The paper considers the task of extracting relations of the type subcellular-localization(Protein, Subcellular-Structure), which represent the various subcellular structures in which proteins may be present. The authors first construct a labelled set of instances which contain the target relation. This is done by selecting six proteins and querying them in MEDLINE, which is a corpus of abstracts of biomedical journals. The abstracts are then hand-annotated with instances of the target relation, subcellular-localization. This resulted in a total of 33 instances of the target relation. This is used to train a Naive Bayes classifier, and later for evaluating the Distant Supervision trained model.
  
The Distant Supervision model, the authors consider the Yeast Protein Database \cite{hodges1999yeast}, which includes a subcellular-localization field for many proteins, as well as a hyperlink to the reference article in PubMed. Each of these entries, along with the reference is used as a weakkly learned instance for relation extraction. After cleaning up the dataset, the authors generate a set of 336 relational instances described in 633 sentences. The sentences that do not mention the relational instance are treated as negative training examples.
\begin{figure}
\caption{Naive Bayes Model \cite{pham2009unsupervised}}
\label{fig:naive}
\centering
\includegraphics[width=0.5\textwidth]{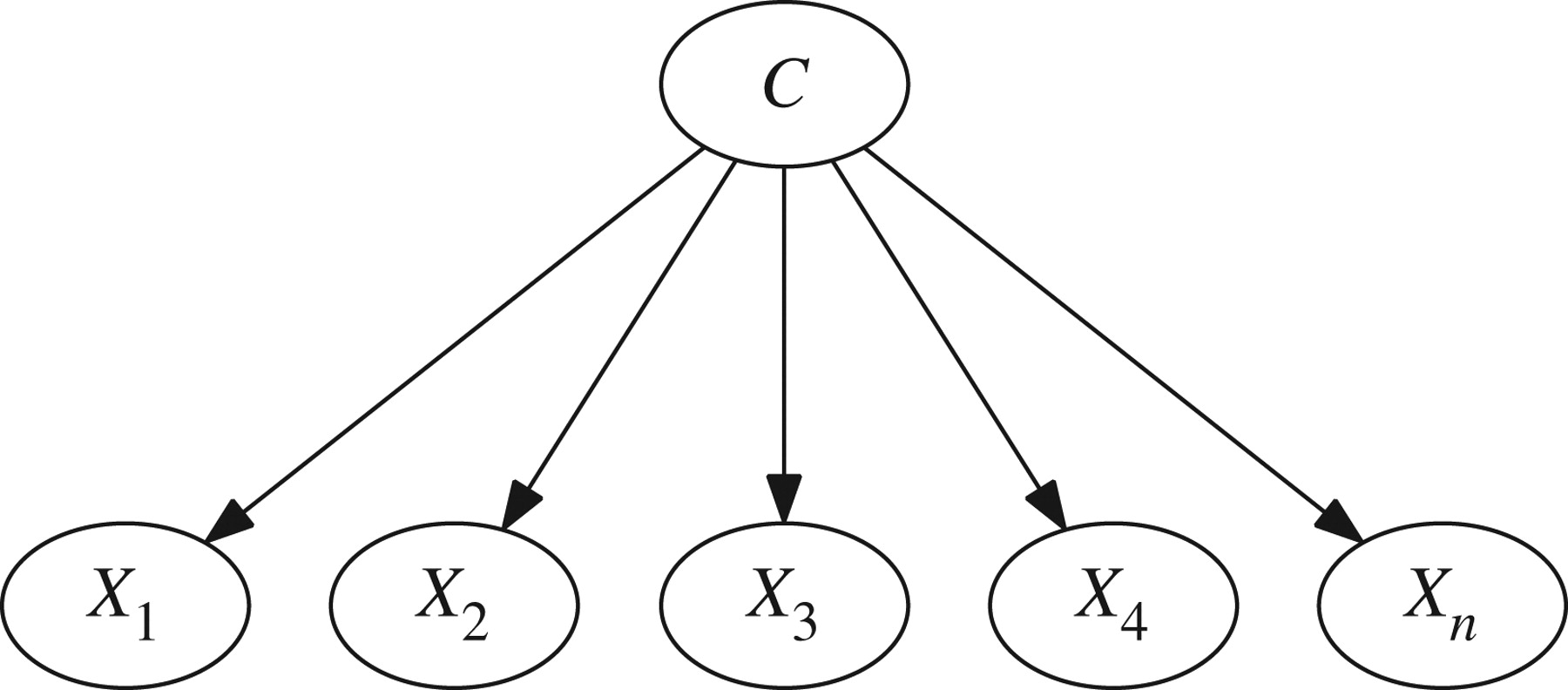}
\end{figure}

The Figure \ref{fig:naive} shows the simple Naive Bayes model. Given a document $d$ of $n$ words, $(w_1, w_2, \cdots, w_n)$, the probability that the document belongs to the class $c_j \in C$ is given by:
\begin{equation}
P(c_j|d) = \frac{P(c_j)\Pi_{i=1}^n P(w_i|c_j)}{P(d)}
\end{equation} 
In this case, there are only two classes, positive and negative, denoting whether the relation is present or not. In order to test whether $r(x,y)$ is a valid relation, all sentences containing both the entities $x$ and $y$ are considered. Further, two approaches are considered. In the first approach, for each sentence, if the predicted class is positive, the extracted relation is returned, else no relation is returned. In the second approach, for each relation a confidence is calculated as follows:
\begin{equation}
confidence = 1 - \Pi_{k=1}^N [1 - P(c = positive|s_k)]
\end{equation}
where $s_k$ is the kth sentence that contains both the entities $x$ and $y$. If the cconfidence is above a threshold, the relation is predicted as positive, and otherwise negative. The authors further evaluate the trained model on the MEDLINE dataset, and observe an increase in performance as compared to the Naive Bayes model trained on the MEDLINE dataset, with cross-validation.

They further add parse tree based features as well to their model, and learn the features through an algorithm similar to FOIL. This leads to a further increase in the accuracy of the model. 

While this model itself is very simple, one of the key contributions of this paper was the novel idea of generating training data through distant supervision and then using the generated data to learn features. 

\subsection{Distant Supervision using Freebase \cite{mintz2009distant}}
This paper describes the key idea of Distant Supervision in a domain independent setting, and uses Freebase as the KB for distant supervision. 

\textbf{The key assumption is that if two entities participate in a relation, a sentence that contains mentions of both the entities, should express the relation.}

The intuition behind this approach is to generate a training set of entity pairs that participate in a given set of relations. Entities are tagged with NER tools in the training step and if a sentence contains two entities known to be an instance of some Freebase relation, features are extracted and added to the feature vector for that relation. However, since sentences could possibly express incorrect relations, a multiclass logistic regression is trained to learn weights on the noisy extracted features.

At the time of testing, given a sentence, entities are identified using NER tools, and every pair of entities that appear together is considered a potential relation. For each entity pairs, features are extracted and the regression classifier predicts a relation name for every entity pair based on the extracted features.

The lexical features extracted are:
\begin{itemize}
\item Sequence of words between entities
\item POS tags of these words
\item A flag indicating the order of appearance 
\item A window of k words to the left of entity 1
\item A window of k words to the right of entity 2
\end{itemize}
The syntactic features are extracted by building a dependency parse tree. The features are:
\begin{itemize}
\item Dependency path between entities
\item A window node for each entity: A window node is a node connected to one of the entities, but is not in the dependency path.
\end{itemize}
All the features in the classifier are used in conjunction. For two features to match, all the subfeatures should match. This is done in order to have high precision features at the cost of low recall.

It is shown that this algorithm is able to extract high precision patterns for a large number of relations, while the held out evaluation shows that combination of lexical and syntactic features performs the best.

\subsection{Multi Instance Learning (MIL) \cite{riedel2010modeling}}
\textbf{This paper argues that the key assumption in current Distant Supervision tecniques}, that each sentence which mentions the two related entities is an expression of the given relation \textbf{is too strong} and leads to noisy patterns that hurts the precision. 

This paper relaxes this assumption to:
\\
\textbf{If two entities participate in a relation, at least one sentence that mentions these two entities might express that relation.}

While this assumption is clearly better, it comes with additional complexity in both testing and training. The paper further suggests a undirected graphical model, which addresses both the tasks of predicting relation between entities and predicting which sentences express this relation.

\begin{figure*}[t]
\caption{Factor Graph Model \cite{riedel2010modeling}}
\label{fig:mil}
\centering
\includegraphics[width = 0.8\textwidth]{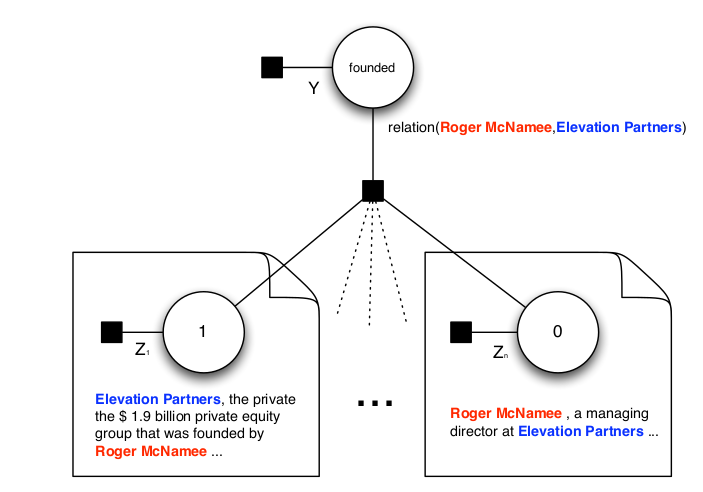}
\end{figure*}

The Figure \ref{fig:mil} \cite{riedel2010modeling} shows an instance for the Factor Graph model. Consider the example shown in the figure. The KB contains the relation founded(Roger McNamee, Elevation Partners). Consider the sentence ``Elevation Partners , the 1.9 billion private equity group that was founded by Roger McNamee ...". This sentence contains mentions of both the entities, as well as expresses the correct relation. Now consider the sentence ``Roger McNamee , a managing director at Elevation Partners , ...". While this sentence does contain mentions of both the entities, it does not express the correct relation. 

The model uses two types of hidden variables. Given a pair of entities, $S$ and $D$, that appear together in at least sentence, a variable $Y$ denotes the relation between them if it exists and $NA$ otherwise. In Figure \ref{fig:mil}, Y is set to $founded$. Further, for the ith sentence that mentions both the entities, the authors define a boolean relation mention variable $Z_i$, which is true iff the ith sentence is indeed mentioning the relation $Y$. The entity mentions in the ith sentence are refered to as $S_i$ and $D_i$ respectively. Additional information about the ith sentence is stored in the observed variable $x_i$. This is collected across all sentences that mention the entities to get the vector $x$. Further, $Z$ is used to denote the state of all mention candidates. The conditional distribution is given by:
\begin{equation}
P(Y=y, Z=z|x) = \frac{\Phi^r(y)\Phi^{join}(y,z,x)\prod_i\Phi^m(z_i, x_i)}{Z_x}
\end{equation}
where $\Phi^r$ denotes the bias of the model towards a relation type y, and is defined as $\Phi^r = exp(\theta^r_y)$. The function $\Phi^m$ is defined as a function over $x_i$ as:
\begin{equation}
\Phi^m(z_i, x_i) = exp(\sum_j \theta_j^m \phi_j^m(z_i, x_i))
\end{equation}
The feature functions $\phi_j^m(z_i, x_i)$ are those defined above as lexical and syntactic features, while the factor $\Phi^join(y,z,x)$ is defined as:
\begin{equation}
\Phi^{join}(y,z,x) = exp(\sum_j \theta_{j,y}^{join} \phi_j^{join}(z, x))
\end{equation}
where $\phi_j^{join}(z, x)$ is defined as:
\begin{equation}
\phi_j^{join}(z, x) = 1  iff \exists i: z_i = 1 \& \phi_j^{m}(z_i, x_i) = 1  
\end{equation}
The feature $\phi_j^{join}$ denotes whether the feature $\phi_j^m$ is active for any of the relation mentions which are correct w.r.t $Y$.

Inference in this setting is primarily performed by Gibbs sampling. Learning on the other hand is done by SampleRank, which is a ranked based learning framework. Since each step of inference is usually the bottleneck in learning, SampleRank has been shown to be efficient in training for models in which inference is intractable.\\
\\
\\Finally, in this setting, an error reduction of 31\% is observed in the NY Times dataset.
\subsection{MultiR \cite{hoffmann2011knowledge}}
\begin{figure*}[t]
\caption{(a) Plate model of the network (b) an example network for entities Steve Jobs, Apple \cite{hoffmann2011knowledge}}
\label{fig:hoff}
\centering
\includegraphics[width = \textwidth]{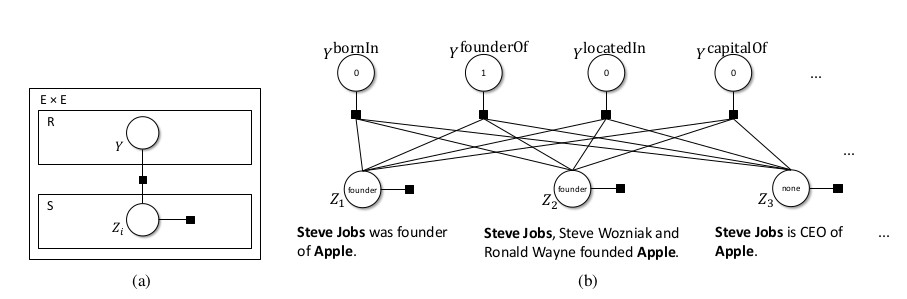}
\end{figure*}

One of the issues in the model by \cite{riedel2010modeling} is that it does not allow relations to overlap, i.e. for any pair of entities $e_1$ and $e_2$, there cannot exist two facts $r(e_1, e_2)$ and $q(e_1, e_2)$. This paper firther relaxes this assumption by constructing a graphical model as shown in Figure \ref{fig:hoff}.

The other obvious change is in the  conditional extraction model, which is now given by:
\begin{dmath}
P(Y=y, Z=z|x) = \\ \frac{1}{Z_x}\prod_r \Phi^{join}(y^r, z) \prod_i \Phi^{extract}(z_i, x_i) 
\end{dmath}
where the functions are the same as those defined above.

Learning in this setting is done based on approximations which lead to Perceptron-style additive parameter updates \cite{collins2002discriminative}. The first approximation is to do online learning instead of the full optimization. The second approximation is to replace expectations with maximizations (Viterbi approximation).

It was experimentally shown that this model has a better precision and recall as compared to the model proposed by \cite{riedel2010modeling}.

\subsection{Multi-Instance Multi-Label (MIML) \cite{surdeanu2012multi}}
\begin{figure}
\caption{MIML plate model \cite{surdeanu2012multi}}
\label{fig:miml}
\centering
\includegraphics[width = 0.3\textwidth]{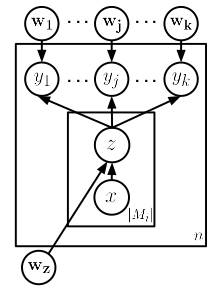}
\end{figure}

This paper also tries to relax the assumption that each relations between entities can overlap. The plate model is shown in Figure \ref{fig:miml}.
The model structure is very similar to that proposed by \cite{hoffmann2011knowledge}, however training is done by Expectation Maximization (EM). In the E step, latent mention labels are assigned using the current model and in the M step, the model is retrained to maximize the log likelihood of the data using the current assignments.

It was observed that while this model performs better than the one proposed by  \cite{riedel2010modeling}, it does not perform as good as \cite{hoffmann2011knowledge}, especially around extremities. The authors however claim that this model is more stable as it yields a smoother curve. 

\subsection{MIML-semi \cite{min2013distant}}
\begin{figure}
\caption{MIML-semi plate model \cite{min2013distant}}
\label{fig:mimlsemi}
\centering
\includegraphics[width = 0.2\textwidth]{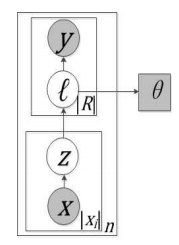}
\end{figure}

This paper builds on the work by \cite{surdeanu2012multi} and shows that due to the incomplete nature of KBs, a significant number of negative examples generated are actually false negatives. They correct this problem by modelling the bag-level noise caused by incomplete KBs, in addition to modelling instance level noise using the MIML model. The plate diagram is shown in Figure \ref{fig:mimlsemi}.
The input to the model is a list of n bags with labels Positive (P) or unlabelled (U) for each relation r. The model adds another set of latent variables, $l$, which models the true bag level labels, to combine the labels y and the MIML layers. Training is done by an EM algorithm similar to that proposed by \cite{surdeanu2012multi}.

This model achieves a better performance than both MIML as well as MultiR on the KBP dataset.

Another interesting approach in modelling missing data is presented in \cite{ritter2013modeling}, which could not be studied due to lack of time.
An interesting aspect of Distant Supervision was also presented in \cite{bunescu2007learning}, which primarily addresses the task of Multi Instance Learning as a supervised learning problem, solved by SVMs.
A summary of the distant supervision techniques along with their key contributions is presented in Table \ref{table:summary}.

\begin{table}[]
\centering
\caption{Summary of Distant Supervision Techniques}
\label{table:summary}
\begin{tabular}{|l|l|l|}
\hline
    & Model                                                              & \begin{tabular}[c]{@{}l@{}}Key Assumptions/\\ Contributions\end{tabular}                                                                                    \\ \hline
4.2 & \begin{tabular}[c]{@{}l@{}}Distant \\ Supervision\end{tabular}     & \begin{tabular}[c]{@{}l@{}}If r($e_1$, $e_2$) $\in$ KB, \\ then $\forall$ s such that\\ $e_1, e_2 \in sentence s$,\\ s expresses the relation r\end{tabular} \\ \hline
4.3 & \begin{tabular}[c]{@{}l@{}}Multi Instance \\ Learning\end{tabular} & \begin{tabular}[c]{@{}l@{}}If r($e_1$, $e_2$) $\in$ KB, \\ $\exists$ s such that\\ $e_1, e_2 \in sentence s$, and\\ s expresses the relation r\end{tabular} \\ \hline
4.4 & MultiR                                                             & \begin{tabular}[c]{@{}l@{}}Allows overlapping \\ relations for entities\end{tabular}                                                                        \\ \hline
4.5 & MIML                                                               & \begin{tabular}[c]{@{}l@{}}Allows overlapping \\ relations for entities\end{tabular}                                                                        \\ \hline
4.6 & MIML-semi                                                          & \begin{tabular}[c]{@{}l@{}}Addresses false negatives \\ generated due to\\  incomplete KB\end{tabular}                                                      \\ \hline
\end{tabular}
\end{table}

\section{Conclusions}
Relation Extraction is a particularly well studied problem in the NLP domain. In this paper, we focussed on one of the technique of Distant Supervision, which utilises a large Knowledge Base, such as Freebase to generate training examples for relational extraction. We reviewed some of the models in Distant Supervision, and presented a summary of some of the key contributions of each model.

\newpage
\bibliography{ref}

\begin{thebibliography}{}
\expandafter\ifx\csname natexlab\endcsname\relax\def\natexlab#1{#1}\fi

\bibitem[{Agichtein and Gravano(2000)}]{agichtein2000snowball}
Eugene Agichtein and Luis Gravano. 2000.
\newblock Snowball: Extracting relations from large plain-text collections.
\newblock In {\em Proceedings of the fifth ACM conference on Digital
  libraries\/}. ACM.

\bibitem[{Bollacker et~al.(2008)Bollacker, Evans, Paritosh, Sturge, and
  Taylor}]{bollacker2008freebase}
Kurt Bollacker, Colin Evans, Praveen Paritosh, Tim Sturge, and Jamie Taylor.
  2008.
\newblock Freebase: a collaboratively created graph database for structuring
  human knowledge.
\newblock In {\em Proceedings of the 2008 ACM SIGMOD international conference
  on Management of data\/}. AcM.

\bibitem[{Brin(1998)}]{brin1998extracting}
Sergey Brin. 1998.
\newblock Extracting patterns and relations from the world wide web.
\newblock In {\em International Workshop on The World Wide Web and
  Databases\/}. Springer.

\bibitem[{Bunescu and Mooney(2007)}]{bunescu2007learning}
Razvan Bunescu and Raymond Mooney. 2007.
\newblock Learning to extract relations from the web using minimal supervision.
\newblock In {\em ACL\/}.

\bibitem[{Bunescu and Mooney(2005)}]{bunescu2005shortest}
Razvan~C Bunescu and Raymond~J Mooney. 2005.
\newblock A shortest path dependency kernel for relation extraction.
\newblock In {\em Proceedings of the conference on human language technology
  and empirical methods in natural language processing\/}. Association for
  Computational Linguistics.

\bibitem[{Collins(2002)}]{collins2002discriminative}
Michael Collins. 2002.
\newblock Discriminative training methods for hidden markov models: Theory and
  experiments with perceptron algorithms.
\newblock In {\em Proceedings of the ACL-02 conference on Empirical methods in
  natural language processing-Volume 10\/}. Association for Computational
  Linguistics.

\bibitem[{Craven et~al.(1999)Craven, Kumlien et~al.}]{craven1999constructing}
Mark Craven, Johan Kumlien, et~al. 1999.
\newblock Constructing biological knowledge bases by extracting information
  from text sources.
\newblock In {\em ISMB\/}.

\bibitem[{Culotta and Sorensen(2004)}]{culotta2004dependency}
Aron Culotta and Jeffrey Sorensen. 2004.
\newblock Dependency tree kernels for relation extraction.
\newblock In {\em ACL\/}. Association for Computational Linguistics.

\bibitem[{Fader et~al.(2011)Fader, Soderland, and
  Etzioni}]{fader2011identifying}
Anthony Fader, Stephen Soderland, and Oren Etzioni. 2011.
\newblock Identifying relations for open information extraction.
\newblock In {\em EMNLP\/}. Association for Computational Linguistics.

\bibitem[{Hodges et~al.(1999)Hodges, McKee, Davis, Payne, and
  Garrels}]{hodges1999yeast}
Peter~E Hodges, Andrew~HZ McKee, Brian~P Davis, William~E Payne, and James~I
  Garrels. 1999.
\newblock The yeast proteome database (ypd): a model for the organization and
  presentation of genome-wide functional data.
\newblock {\em Nucleic Acids Research\/} .

\bibitem[{Hoffmann et~al.(2011)Hoffmann, Zhang, Ling, Zettlemoyer, and
  Weld}]{hoffmann2011knowledge}
Raphael Hoffmann, Congle Zhang, Xiao Ling, Luke Zettlemoyer, and Daniel~S Weld.
  2011.
\newblock Knowledge-based weak supervision for information extraction of
  overlapping relations.
\newblock In {\em Proceedings of the 49th Annual Meeting of the Association for
  Computational Linguistics: Human Language Technologies-Volume 1\/}.
  Association for Computational Linguistics.

\bibitem[{Jenssen et~al.(2001)Jenssen, L{\ae}greid, Komorowski, and
  Hovig}]{jenssen2001literature}
Tor-Kristian Jenssen, Astrid L{\ae}greid, Jan Komorowski, and Eivind Hovig.
  2001.
\newblock A literature network of human genes for high-throughput analysis of
  gene expression.
\newblock {\em Nature genetics\/} .

\bibitem[{Ji et~al.(2010)Ji, Grishman, Dang, Griffitt, and
  Ellis}]{ji2010overview}
Heng Ji, Ralph Grishman, Hoa~Trang Dang, Kira Griffitt, and Joe Ellis. 2010.
\newblock Overview of the tac 2010 knowledge base population track.
\newblock In {\em Third Text Analysis Conference (TAC 2010)\/}.

\bibitem[{Mikolov et~al.(2013)Mikolov, Sutskever, Chen, Corrado, and
  Dean}]{mikolov2013distributed}
Tomas Mikolov, Ilya Sutskever, Kai Chen, Greg~S Corrado, and Jeff Dean. 2013.
\newblock Distributed representations of words and phrases and their
  compositionality.
\newblock In {\em NIPS\/}.

\bibitem[{Min et~al.(2013)Min, Grishman, Wan, Wang, and
  Gondek}]{min2013distant}
Bonan Min, Ralph Grishman, Li~Wan, Chang Wang, and David Gondek. 2013.
\newblock Distant supervision for relation extraction with an incomplete
  knowledge base.
\newblock In {\em HLT-NAACL\/}. pages 777--782.

\bibitem[{Mintz et~al.(2009)Mintz, Bills, Snow, and
  Jurafsky}]{mintz2009distant}
Mike Mintz, Steven Bills, Rion Snow, and Dan Jurafsky. 2009.
\newblock Distant supervision for relation extraction without labeled data.
\newblock In {\em Proceedings of the Joint Conference of the 47th Annual
  Meeting of the ACL and the 4th International Joint Conference on Natural
  Language Processing of the AFNLP: Volume 2-Volume 2\/}. Association for
  Computational Linguistics.

\bibitem[{Owczarzak and Dang(2011)}]{owczarzak2011overview}
Karolina Owczarzak and Hoa~Trang Dang. 2011.
\newblock Overview of the tac 2011 summarization track: Guided task and aesop
  task.
\newblock In {\em Proceedings of the Text Analysis Conference (TAC 2011),
  Gaithersburg, Maryland, USA, November\/}.

\bibitem[{Pham and Ruz(2009)}]{pham2009unsupervised}
Duc~Truong Pham and Gonzalo~A Ruz. 2009.
\newblock Unsupervised training of bayesian networks for data clustering.
\newblock In {\em Proceedings of the Royal Society of London A: Mathematical,
  Physical and Engineering Sciences\/}. The Royal Society.

\bibitem[{Riedel et~al.(2010)Riedel, Yao, and McCallum}]{riedel2010modeling}
Sebastian Riedel, Limin Yao, and Andrew McCallum. 2010.
\newblock Modeling relations and their mentions without labeled text.
\newblock In {\em Joint European Conference on Machine Learning and Knowledge
  Discovery in Databases\/}. Springer.

\bibitem[{Ritter et~al.(2013)Ritter, Zettlemoyer, Etzioni
  et~al.}]{ritter2013modeling}
Alan Ritter, Luke Zettlemoyer, Oren Etzioni, et~al. 2013.
\newblock Modeling missing data in distant supervision for information
  extraction.
\newblock {\em Transactions of the Association for Computational Linguistics\/}
  .

\bibitem[{Sarawagi et~al.(2004)Sarawagi, Cohen et~al.}]{sarawagi2004semi}
Sunita Sarawagi, William~W Cohen, et~al. 2004.
\newblock Semi-markov conditional random fields for information extraction.
\newblock In {\em NIPS\/}.

\bibitem[{Schmitz et~al.(2012)Schmitz, Bart, Soderland, Etzioni
  et~al.}]{schmitz2012open}
Michael Schmitz, Robert Bart, Stephen Soderland, Oren Etzioni, et~al. 2012.
\newblock Open language learning for information extraction.
\newblock In {\em Proceedings of the 2012 Joint Conference on Empirical Methods
  in Natural Language Processing and Computational Natural Language
  Learning\/}. Association for Computational Linguistics.

\bibitem[{Surdeanu et~al.(2012)Surdeanu, Tibshirani, Nallapati, and
  Manning}]{surdeanu2012multi}
Mihai Surdeanu, Julie Tibshirani, Ramesh Nallapati, and Christopher~D Manning.
  2012.
\newblock Multi-instance multi-label learning for relation extraction.
\newblock In {\em Proceedings of the 2012 joint conference on empirical methods
  in natural language processing and computational natural language
  learning\/}. Association for Computational Linguistics.

\bibitem[{Zeng et~al.(2014)Zeng, Liu, Lai, Zhou, Zhao
  et~al.}]{zeng2014relation}
Daojian Zeng, Kang Liu, Siwei Lai, Guangyou Zhou, Jun Zhao, et~al. 2014.
\newblock Relation classification via convolutional deep neural network.
\newblock In {\em COLING\/}.

\end{thebibliography}
\bibliographystyle{acl_natbib}

\end{document}